\documentclass{article} 
\usepackage{collas2022_conference,times}

\usepackage{amsmath,amsfonts,bm}

\def\eqref#1{equation~\ref{#1}}

\def\1{\bm{1}}

\DeclareMathAlphabet{\mathsfit}{\encodingdefault}{\sfdefault}{m}{sl}
\SetMathAlphabet{\mathsfit}{bold}{\encodingdefault}{\sfdefault}{bx}{n}

\usepackage{hyperref}
\usepackage{xspace}
\usepackage{listings}
\newcommand{\tella}{\textsc{tella}\xspace}
\usepackage{caption}
\usepackage{subcaption}

\lstdefinestyle{mystyle}{
  basicstyle=\ttfamily\footnotesize,
  breakatwhitespace=false,         
  breaklines=true,                 
  captionpos=b,                    
  keepspaces=true,                 
  numbersep=5pt,                  
  showspaces=false,                
  showstringspaces=false,
  tabsize=2
}

\hypersetup{
    colorlinks=true,
    linkcolor=red,
    filecolor=magenta,      
    urlcolor=blue,
    citecolor=purple,
    pdftitle={Overleaf Example},
    pdfpagemode=FullScreen,
    }

\usepackage[textsize=tiny]{todonotes}
\usepackage{comment}

\title{Continual Reinforcement Learning with \tella}

\author{
Neil Fendley$^1$   \\
\And
Cash Costello$^1$   \\
\And
Eric Nguyen$^1$   \\
\And
Gino Perrotta$^1$  \\
\And 
Corey Lowman$^1$  \\
\And 
$^1$Johns Hopkins University Applied Physics Lab \\
\texttt{First.Last@jhuapl.edu} \\
United States\\
}

\collasfinalcopy 

\begin{document}

\maketitle

\begin{abstract}
Training reinforcement learning agents that continually learn across multiple environments is a challenging problem.
This is made more difficult by a lack of reproducible experiments and standard metrics for comparing different continual learning approaches. 
To address this, we present \tella, a tool for the \textbf{t}est and \textbf{e}valuation of \textbf{l}ifelong \textbf{l}earning \textbf{a}gents.
\tella provides specified, reproducible curricula to lifelong learning agents while logging detailed data for evaluation and standardized analysis.
Researchers can define and share their own curricula over various learning environments or run against a curriculum created under the DARPA Lifelong Learning Machines (L2M) Program.
\end{abstract}

\section{Introduction}
\label{intro}
In the last decade, reinforcement learning (RL) with deep neural networks has been successfully applied in a wide variety of domains \citep{arulkumaran_drlsurvey_2017}. In typical RL scenarios, the RL agent learns a single task, defined as a single Partially Observable  Markov Decision Process (POMDP). However, it is more realistic that agents encounter nonstationary tasks or may interact with multiple tasks. For example, an agent controlling an unmanned ground vehicle (UGV) may encounter a novel weather condition (e.g., icy road at dusk) and deal with new states (lighting at dusk or radar reflectance with ice) and new transitions (steering actions have a different effect on an icy road compared to a dry road). Consequently, the field of \textit{continual} reinforcement learning has received increased research interest in the past few years \citep{khetarpal_rlreview_2020, parisi_continual_2019}. Typically, \textit{continual learning} indicates that the agent encounters a sequence of tasks and has to master each task  while maintaining performance on the previously-learned tasks; and \textit{lifelong learning} is indicates the agent is able to use the knowledge from previously-learned tasks to become a better learner over time (learn new tasks better and faster). For purposes of this paper, we use \textit{continual learning} as the more general term encompassing both kinds of abilities.

The assessment of continual learning poses a challenge beyond that of developing the algorithm. Reinforcement learning is extremely variable \citep{chan_reliability_2020}\citep{DRL_at_the_Edge}, and careful evaluation is needed to assess whether an agent can reliably and robustly learn a given task. The challenge is compounded when the agent has to learn multiple tasks, as the results depend on the precise characteristics of the tasks, their presentation sequence, the number of runs (i.e., the number of times the the sequence is repeated), and so on. This variability implies that, in order to compare different RL agents, it is essential to exercise the agents on carefully controlled and replicable scenarios. 

To address the above use case, we introduce an environment, \tella \footnote{\tella is available at \url{https://github.com/lifelong-learning-systems/tella}.}, \textbf{t}est and \textbf{e}valuation of \textbf{l}ifelong \textbf{l}earning \textbf{a}gents, that provides a standardized evaluation framework for episodic lifelong reinforcement learning. \tella is a framework for the specification and execution of experiments in lifelong reinforcement learning. It interfaces with existing collections of environments using the OpenAI Gym API \citep{brockman2016openai}, provides flexible, standard structures for the definition of RL agents and curricula, and enables calculation of standard metrics for continual and lifelong learning. Using \tella, researchers can create and share benchmarks by providing curricula as standardized Python code which can be used to repeatably train and evaluate agents. This allows for replicable results and makes assessing progress in continual learning algorithms feasible. Note that \tella itself is \textit{not} a common benchmark problem (although several curricula are included in the package). It is a tool to aid in both the development and use of user-specified benchmarks. 

\tella was developed to standardize evaluation and comparison of agents in the DARPA Lifelong Learning Machines (L2M) program and has been successfully used to evaluate a variety of different continual RL agent architectures on a common set of benchmark tasks and scenarios.

\section{Related Work}

There have been several efforts to develop environments to facilitate such assessment. One line of effort is to define benchmark problems, such as OpenAI's Procgen Benchmark \citep{openAIprocgen}, Meta-World's multi-task environments \citep{yu2019meta}, the Arcade Learning Environment's 50+ Atari games \citep{bellemare13arcade}, or the specific subset of six Atari games specified by \citet{pmlr-v80-schwarz18a}. However, benchmark problems are not sufficient for standardized evaluation, since experiments using these benchmarks may still differ significantly in terms of the order of tasks and the amount of interaction the agent is allowed to take in each task (i.e., the amount of data collected for learning). Another line of effort has to been establish ground truth with standard metrics: ShinRL \citep{kitamura_shinrl_2021} defines tasks with known optimal solutions (Q functions); CORA \citep{powers_cora_2021} includes a set of benchmark tasks, standard scenarios and metrics. However, this line of effort does not address an extensible way to assess black-box RL agents, using reproducible experiments with parameterized tasks, customizable scenarios (curricula), multiple runs, logging and continual learning metrics.  

We note that the recently-introduced Avalanche RL \citep{lucchesi_avalancherl_2022} is a parallel effort with similar goals, though there are differences of emphasis. In particular, \tella allows flexible evaluation scenarios and schedules, and is agnostic to the internal structure and learning procedure of the agent (such as optimization or number of models) and is thus better able to accommodate multi-model RL architectures.

\section{Design Principles}
To succeed as a benchmarking tool for continual reinforcement learning, \tella meets several conceptual requirements: \tella must be compatible with curricula and agents authored independently, it must be configurable to permit varied agent and environment design, and it must execute experiments in a repeatable and efficiently scalable manner.


\subsection{Compatible}
\tella defines a standard structure for agents and curricula which is designed for compatibility even when the agent and curriculum are developed separately. 
This standardization is accomplished by requiring that agents implement a specific interface that can be normally be achieved with a simple wrapper around the original agent code. Curricula are specified as sequences of agent-environment interactions with all environments required to use the OpenAI Gym API \citep{brockman2016openai} and share action and observation spaces. \tella does not restrict the contents of an agent implementation, but the agent must respond to and provide feedback for the curriculum event callbacks.
This design ensures that \tella's experiment execution method has the necessary connections to drive and record the interactions between agent and curriculum.

\subsection{Configurable}
\tella aims for minimal restrictions on curricula and agents. Curricula can be defined for any environments which conform to the OpenAI Gym API. Interactions between agent and environment are managed by a method within \tella, rather than by the agent itself. While this re-frames some of the way an agent receives data, it does not fundamentally change the algorithm. To verify this, an example agent is provided with the \tella library which organizes the contents of the Stable Baselines3 \citep{stable-baselines3} Proximal Policy Optimization (PPO) agent \citep{PPO} into the \tella callbacks. The original and reorganized agents perform identically.
\subsection{Repeatable}
Experiments in \tella are constructed to permit exact repeatability when possible. Random number generator (RNG) seed values are provided and logged for the agent and curriculum. Example curricula and agents are included to demonstrate the use of these seeds.

\subsection{Scalable}
\tella permits experiment acceleration by environment vectorization. Vectorization allows for an agent to interact with multiple environments in parallel. In order to scale to curricula containing thousands of environments, \tella delays the construction of the environments within a curriculum until they are needed by an agent. Environments are closed when they are not in use. This is useful for environments such as Unity and Starcraft which may have significant memory and runtime requirements.


\begin{figure}
\hspace*{-0.6cm}\captionbox{\tella pipeline diagram \label{fig:tella_diag}}
[.5\textwidth]{\includegraphics[width=.4\textwidth]{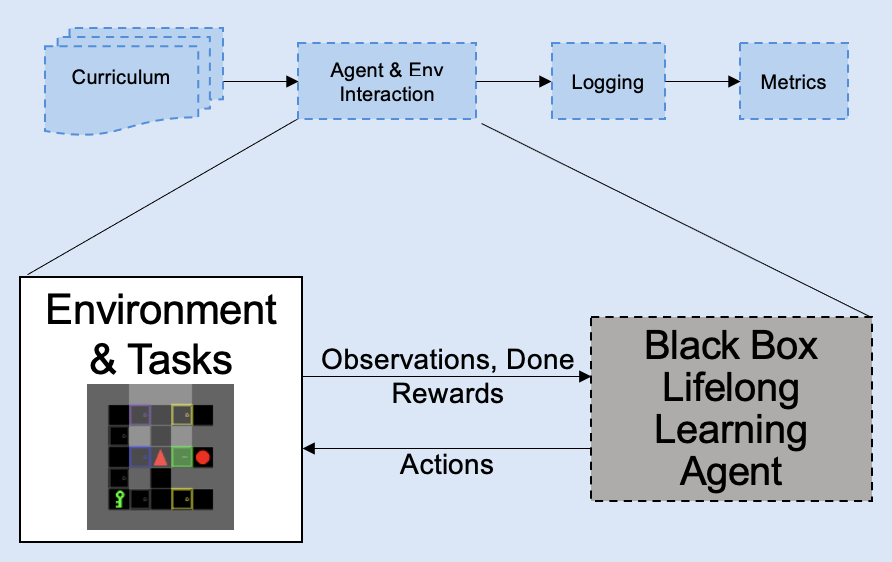}}%
\captionbox{Performance of an agent over a curriculum of 6 custom Minigrid tasks generated with \tella and L2Metrics package
\label{fig:example_results}}
[.5\textwidth]{\includegraphics[width=.5\textwidth]{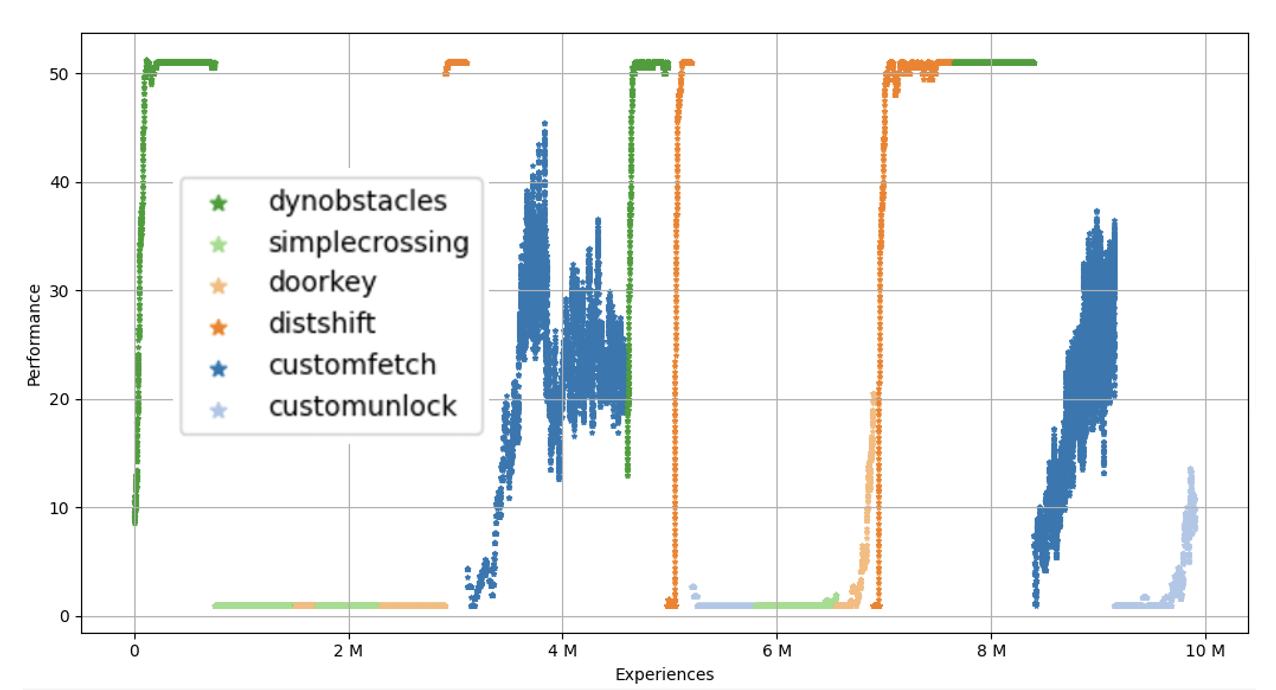}}
\end{figure}

\section{\tella Design}
\label{sec:design}
\subsection{Curricula}

\begin{center}
\begin{tabular}{ |c|c| }
 \hline
 \textbf{Term} & \textbf{Definition} \\
 \hline \hline
 Curriculum & Sequence of Blocks \\ 
 \hline
 Block & Sequence of Task Blocks \\  
 \hline
 Task Block & Sequence of Task Variants \\
 \hline
 Task Variant & An RL environment \\ 
 \hline
\end{tabular}
\end{center}

In order to standardize the training and evaluation of reinforcement learning methods, we define a curriculum structure. A curriculum consists of training blocks and evaluation blocks in sequence. Each block is a collection of tasks and associated parameters within each task. Figure \ref{fig:example_results} shows a curriculum with 6 different tasks. Each of these tasks and associated parameters, called task variants, define an environment to be interacted with by the agent. 

For each interaction with the environment, the agent will be given the observations and must return the actions to use. The environment will then return a reward, the new observation of the environment, and a done signal, shown in Figure \ref{fig:tella_diag}. A done signal indicates an episode is finished and that the environment must be reset to the beginning.

A reinforcement learning agent will increase its performance the more data it has to train on, so comparing across agents requires us to regulate interactions with the environment in some way. We do this by limiting exposure to each task variant, either in the number of steps an agent is allowed to perform or the number of episodes that can be completed. We provide support for both step and episode limiting because which is more relevant can be application dependent. By limiting the interactions with the environment, we can calculate an agent's sample efficiency, which is the amount of samples required to reach the agent's measured performance.
\subsection{Creating Curricula}
In order to create a curriculum one must define the sequence of training and evaluation blocks. The block type is indicated to the agent, and rewards are hidden from the agent during evaluation. Each block represents a sequence of task variants. If the order of task variants within a block is chosen randomly, the seed for repeatable RNG will be recorded. Each task variant defines an environment, any specific parameters needed to repeatably initialize that environment, and a limit on the agent's interaction as either steps or episodes.

We also provide an InterleavedEvalCurriculum, which interleaves the same evaluation block between each training block. This enables users to specify the evaluation block once rather than duplicating it.



\subsection{\tella Agent}

The agent class requires implementing two methods, and a number of optional events that it can use to organize functionality. The first main method is the `choose\_actions' method which receives a list of observations and must return a list of actions. Each observation in the list corresponds to one environment, and, when vectorized environments are used, there will be an observation for each environment. The second main method is the `receive\_transitions' method, which allows the agent to receive the result of applying the actions to the environments. A transition is defined as a tuple of the observation, the action that was taken, the reward for that action, whether the episode is complete, and the next observation after applying the action. `Receive\_transitions' receives a list of transitions, with one transition per environment. This interaction can be summarized as:

\begin{lstlisting}
    actions = agent.choose_actions(observations)
    transitions = vec_env.step(actions)
    agent.receive_transitions(transitions)
\end{lstlisting}

Other events the agent can use are related to events in the curriculum, and \tella automatically calls these methods on the agent at the appropriate time in the curriculum.


\subsection{\tella Experiment}


An experiment in \tella consists of a curriculum, an agent, and a number of times to run the agent through the curriculum. Each single run of the agent through the curriculum is called a lifetime. \tella will automatically determine new seeds for each lifetime in the experiment.

For a single lifetime, \tella will recursively iterate over every block, task block, and task variant, and run the interaction between the task variant's environment and the agent. Once interaction with a task variant is complete, it will move onto the next task variant or step up a level.

The output of a single lifetime is a directory that contains all the episode information from each training and evaluation block. The files logged in the directory can be consumed by a framework such as l2metrics \citep{l2metrics}, to produce lifelong learning metrics.

To make running experiments with an agent easy, \tella includes a function to turn any valid \tella agent into a command line interface. This creates a file that allows you to specify arguments and start experiments from the command line. Alternatively the experiment can be called directly if you know which curriculum you want to run.




\section{Conclusion and Future Work}
Continual reinforcement learning is a research problem that requires an algorithm to learn across dynamic environments. Algorithms in this setting struggle to remember information from previous tasks and to learn new tasks efficiently. To complicate matters, benchmarking continual reinforcement learning algorithms is made more difficult by a lack of standardized procedures. 

To address this difficulty in evaluating continual learning algorithms, \tella standardizes the training and evaluation of episodic continual learning algorithms using curricula. A curriculum is a sequence of training and evaluation tasks with controlled access to training data generated by the environment. By using a curriculum, we can calculate many metrics, including: performance maintenance, to measure catastrophic forgetting; sample efficiency, by limiting the amount of data; and forward transfer, to measure how much an agent is retaining information from previous environments. \tella was developed as part of DARPA's L2M program, and served as a framework for a common evaluation of L2M lifelong learning agents. 

However, there is still future work to be done, both within continual learning and \tella. \tella is focused on the task of episodic learning, as opposed to continuous learning in one dynamic environment. \tella also restricts all access to data in the environment to be pre-specified in the curriculum. This means there is no online control for the agent to adjust the future curriculum. \tella could be extended to allow agents to control the future blocks in the curriculum, which could potentially host active learning or early stopping experiments. 


\section{Acknowledgments}
This effort was funded by the DARPA Lifelong Learning Machines (L2M) Program.  The views, opinions, and/or findings expressed are those of the author(s) and should not be interpreted as representing the official views or policies of the Department of Defense or the U.S. Government.

\bibliography{tella_refs}
\bibliographystyle{collas2022_conference}


\end{document}